\documentclass[11pt]{article}
\usepackage[english]{babel}
\usepackage{amssymb,amsmath,graphicx}
\usepackage{amsthm}
\usepackage{float}
\usepackage{afterpage}
\usepackage{color}
\usepackage{enumitem}
\usepackage{setspace}
\usepackage{algorithmic}
\usepackage{algorithm}
\usepackage[table]{xcolor}
\usepackage{multirow}
\usepackage[numbers]{natbib}
\usepackage{microtype}
\usepackage[hidelinks]{hyperref}
\newcommand{\bftab}{\fontseries{b}\selectfont}
\usepackage[titletoc,title]{appendix}

\newcommand{\cO}{\mathcal{O}}

\title{A simple and effective hybrid genetic search for the job sequencing and tool switching problem}

\begin{document}

\begin{center}

\begin{LARGE}
A Simple and Effective Hybrid Genetic Search for the\vspace*{0.25cm}\linebreak  Job Sequencing and Tool Switching Problem
\end{LARGE}

\vspace*{1cm}

\textbf{Jordana Mecler} \\
Departamento de Inform\'{a}tica, Pontif\'{i}cia Universidade Cat\'{o}lica do Rio de Janeiro (PUC-Rio) \\
\url{jmecler@inf.puc-rio.br} \\
\vspace*{0.35cm}

\textbf{Anand Subramanian} \\
Universidade Federal da Para{\'i}ba, Departamento de Sistemas de Computa{\c c}{\~a}o\\
\url{anand@ci.ufpb.br}\\
\vspace*{0.35cm}

\textbf{Thibaut Vidal} \\
Departamento de Inform\'{a}tica, Pontif\'{i}cia Universidade Cat\'{o}lica do Rio de Janeiro (PUC-Rio) \\
\url{vidalt@inf.puc-rio.br}\\

\vspace*{1cm}

\end{center}

\noindent
\textbf{Abstract.} The job sequencing and tool switching problem (SSP) has been extensively studied in the field of operations research, due to its practical relevance and methodological interest. Given a machine that can load a limited amount of tools simultaneously and a number of jobs that require a subset of the available tools, the SSP seeks a job sequence that minimizes the number of tool switches in the machine. To solve this problem, we propose a simple and efficient hybrid genetic search based on a generic solution representation, a tailored decoding operator, efficient local searches and diversity management techniques. To guide the search, we introduce a secondary objective designed to break ties. These techniques allow to explore structurally different solutions and escape local optima. As shown in our computational experiments on classical benchmark instances, our algorithm significantly outperforms all previous approaches while remaining simple to apprehend and easy to implement. We finally report results on a new set of larger instances to stimulate future research and comparative analyses.

\vspace*{0.5cm}


\vspace*{0.5cm}

\thispagestyle{empty}
\pagenumbering{arabic}

\pagebreak

\section{Introduction}
\label{sec:introduction}

The \textit{job sequencing and tool switching problem} (SSP) considers a set of jobs $J = \{1,...,n\}$, a set of tools $T = \{1,...,m\}$, and a single machine whose magazine can hold up to $C$ tools simultaneously. Each job $j \in J$ requires a subset $T_j$ of the available tools to be performed, where $|T_j| \leq C$. We assume that the tool setup times are identical and that every tool fits into one slot of the machine. In most applications, the total number of tools needed by all the jobs exceeds the machine magazine capacity. Therefore, the machine must switch some tools when finishing a job and starting another one. The goal of the SSP is to find a job sequence and a tool-loading strategy that minimizes the total number of switches.

The SSP arises in many applications, including circuit-board manufacturing \citep{Privault1995}, computer memory management \citep{Ghiani2007}, and pharmaceutical packaging~\citep{Torsten2014}. In the first application, the jobs represent circuits which need to be placed on a board and the tools are the surface-mount component (SMC) feeders. In the second application, the jobs represent tasks which must be processed and the tools are the pages (memory fragments) that need to be transferred from the slow memory to the fast memory. In the last application, the jobs represent patient medicine boxes, and the tools are the pills. 

The SSP can be generally decomposed into a \emph{sequencing} problem which aims to find the best job sequence, and a \emph{tooling problem} which seeks the best tool loading policy for a fixed sequence. For a fixed job sequence, the tooling problem can be solved in polynomial time in $\cO(n m)$ with the 
``Keep Tools Needed Soonest'' (KTNS) policy \citep{TangAndDenardo1988}. In contrast, for a free job sequence, the SSP is known to be $\mathcal{NP}$-hard \citep{Crama1994}. 

Due to its rich structure and practical interest, the SSP has been the focus of extensive research, and most of the classical metaheuristic frameworks (e.g., tabu search, iterated local search, and genetic algorithms) have been adapted to this problem. However, the discrete nature of the objective (number of tool switches) and the problem symmetries tend to diminish the effectiveness of local searches, and most of the best methods tend to be complex and over-engineered. As a consequence, notable methodological breakthroughs remain achievable, even for medium-scale problems with a few dozens of jobs. Finally, the problem decision sets lend themselves to an effective application of solution representative- and decoder-based algorithms, consisting in searching the space of the permutations and systematically applying a decoder (the KTNS policy) to evaluate solution costs. Such a decision-set problem decomposition has been instrumental in designing efficient metaheuristics for a variety of permutation and set based problems \citep[see, e.g.,][]{Gribel2019,Toffolo2019,Vidal2017b,Vidal2014b,Vidal2014}.

In this article, we present new methodological advances for the SSP and pursue the study of decision set decomposition-based metaheuristics. Inspired by the success of hybrid genetic searches (HGS) on vehicle routing problems ---i.e., genetic algorithms with local search and population diversity management \cite{Vidal2012,Vidal2014}--- we use and adapt this methodological paradigm with search operators tailored for the SSP. To perform a controlled exploration of solutions with identical cost, we also exploit a secondary objective which favors small 0-blocks to guide the search towards solution improvements. The main contributions of this article are fourfold:
\begin{itemize}[nosep]
    \item We introduce an efficient HGS for the SSP. This method exploits a simple permutation-based solution representation and measures solution quality in terms of cost and contribution to the population diversity during parents and survivors selections.
    \item We introduce a secondary objective which promotes short 0-blocks as a means to progress towards solutions with fewer switches.
	\item We conduct extensive computational experiments on classical SSP instances to evaluate the performance of our approach and measure the contribution of its principal components. These experiments show that the proposed method performs remarkably well in relation to previous approaches and that its main strategies (secondary objective and diversity management) are critical for a good performance.
	\item We finally report additional results on a set of larger-scale instances to stimulate future research and comparative analyses.
\end{itemize}

The remainder of the paper is organized as follows. Section~\ref{sec:problemstatement} reviews the literature, while Section~\ref{sec:proposedmethodology} describes the proposed algorithm. Section~\ref{sec:computationalexperiments} presents our computational analyses, and Section~\ref{sec:conclusion} concludes.

\section{Related Studies}
\label{sec:problemstatement}

The SSP was formally proposed in the late 1980s by \citet{TangAndDenardo1988}. Along with the problem, the authors introduced the KTNS policy, which determines the optimal number of tool switches in polynomial time given a predefined job sequence. The KTNS policy is a fundamental example of a greedy algorithm, which is also commonly used for interval scheduling, coloring and caching problems \citep{Bouzina1996,Carlisle1995}.
Later, \citet{Gray1993} discussed decision models for tooling management, including the SSP, while \citet{Crama1994} proved that the job sequencing part of the problem is $\mathcal{NP}$-hard. A model using a network flow approach for the nonuniform case was proposed by \citet{Privault1995}. \citet{Crama1997} discussed optimization models and solutions for production planning and scheduling problems such as the SSP. An empirical study of the SSP in manufacturing companies was conducted by \citet{Shirazi2001}, where the authors concluded that the heuristics available at the time outperformed the methods used within the companies. Reflecting the variety of application cases, several other studies have considered variant of the SSP, e.g., minimizing the tool switching instants \cite{Adjiashvili2015, Konak2007, Tang1988-2}, considering multiple objectives \cite{Keung2001, Salonen2006-2}, parallel machines \cite{Beezao2017, Fathi2002, Zeballos2010}, different tool sizes \cite{Matzliach1998, Tzur2004}, and other features \cite{Avci1996, Furrer2017, Ghrayeb2003, Hertz1996, Raduly-Baka2015, Salonen2006-2, Tzur2004}.

Some mathematical programming algorithms have been proposed for the SSP.
\citet{TangAndDenardo1988} formulated the problem as an integer program (IP). Later on, \citet{Laporte2004}, \citet{Catanzaro2015} and \citet{Ghiani2007} proposed branch-and-cut and branch-and-bound methods based on integer linear programs (ILP). On the other hand, \citet{Bard1988} developed a nonlinear IP and solved it using a dual-based relaxation heuristic, whereas \citet{Ghiani2010} modeled the problem as a nonlinear least-cost Hamiltonian cycle problem and developed a branch-and-cut algorithm. \citet{Yanasse2009} proposed an enumeration method based on a partial ordering of jobs. Finally, \citet{Crama1999} studied the worst-case performance of approximation algorithms for various tool management problems, including the SSP.

Other SSP studies have been principally focused on heuristics and metaheuristics. \citet{TangAndDenardo1988} presented a job scheduling heuristic to find a short Hamiltonian path on a graph, where the nodes represent jobs and the edge weights represent the minimum number of tool switches obtained by processing two jobs consecutively. \citet{Salonen2006-2} considered both job-grouping and tool-switch objectives but also reported heuristic results for the classical SSP. \citet{Burger2015} developed job-grouping heuristics for the color print scheduling problem, which is an application of the SSP. \citet{Crama1994} proposed heuristics based on the traveling salesman problem (TSP) for a graph representation similar to that of \citet{TangAndDenardo1988}. \citet{Hertz1998} improved on earlier heuristics using methods such as GENIUS \cite{Gendreau1992}. \citet{Privault1995} reduced the tool-switching part of the general nonuniform problem case into a search for a minimum cost flow of maximum value over an acyclic network, and they proposed a set of heuristics for the job sequencing. \citet{Djellab2000} formulated the problem using a hypergraph representation and implemented a constructive method. \citet{Salonen2006} presented a multi-start algorithm, creating the starting points by grouping jobs with similar tools.

A variety of local search-based metaheuristics have been applied to the problem, including tabu search \citep{AlFawzan2003,Konak2008}, iterated local search \citep{Paiva2017}, and beam search \citep{Senne2009, Zhou2005}. Population-based methods have also been successful. Hybrid methods combining genetic algorithms (GA) with other search procedures can be found in \cite{Ahmadi2018, Amaya2008, Amaya2011, Amaya2013, Amaya2012, Amaya2010, Chaves2016}. \citet{Amaya2008, Amaya2013, Amaya2012} combined GA with local search-based procedures such as hill climbing, simulated annealing and tabu search, leading to hybrid methods which are also known under the name of memetic algorithms. \citet{Amaya2011, Amaya2013, Amaya2010} combined genetic algorithms with a multi-agent approach or cross-entropy methods. Finally, \citet{Chaves2016} combined clustering search (CS) with a biased random-key genetic algorithm (BRKGA), while \citet{Ahmadi2018} combined a 2-TSP scheme with a novel learning-based GA.
\autoref{tab:works} summarizes the SSP studies in chronological order, lists their main contributions as well as the origin of the benchmark instances considered.

\begin{table}[htbp]
\centering
\renewcommand{\arraystretch}{1.2}
\caption{Summary of SSP studies}
\footnotesize{}
\setlength{\tabcolsep}{3.5pt}
\begin{tabular}{ l l l l } 
 \hline
 \textbf{Work} & \textbf{Year} & \textbf{Approach} & \textbf{Instances} \\
 \hline
 \citet{TangAndDenardo1988} & 1988 & Exact methods + heuristics & \cite{TangAndDenardo1988} \\ 
 \citet{Bard1988} & 1988 & Exact methods + heuristics & \cite{Bard1988} \\
 \citet{Crama1994} & 1994 & Heuristics & \cite{Crama1994} \\ 
 \citet{Privault1995} & 1995 & Network flow formulation + heuristics & \cite{Privault1995} \\ 
 \citet{Hertz1998} & 1998 & Heuristics & \cite{Hertz1998} \\ 
 \citet{Crama1999} & 1999 & Worst-case analysis & -- \\ 
 \citet{Djellab2000} & 2000 & Heuristics & \cite{Crama1994} \\ 
 \citet{Shirazi2001} & 2001 & Empirical study & Real life \\ 
 \citet{AlFawzan2003} & 2003 & Tabu search & \cite{AlFawzan2003} \\ 
 \citet{Laporte2004} & 2004 & Exact methods & \cite{Hertz1998, Laporte2004} \\ 
 \citet{Zhou2005} & 2005 & Beam search & \cite{Bard1988} \\ 
 \citet{Salonen2006} & 2006 & Heuristics & \cite{Crama1994} \\
 \citet{Salonen2006-2} & 2006 & Exact methods + heuristics & \cite{Salonen2006-2} \\
 \citet{Ghiani2007} & 2007 & Exact methods & \cite{Hertz1998, Laporte2004} \\ 
 \citet{Amaya2008} & 2008 & Memetic algorithms & \cite{Amaya2008} \\
 \citet{Konak2008} & 2008 & Tabu search & \cite{Konak2008} \\ 
 \citet{Senne2009} & 2009 & Beam search & \cite{Senne2009} \\ 
 \citet{Yanasse2009} & 2009 & Exact methods & \cite{Yanasse2009} \\
 \citet{Ghiani2010} & 2010 & Exact methods & \cite{Ghiani2010} \\ 
 \citet{Amaya2010} & 2010 & Hybrid cooperative & \cite{Amaya2010} \\
 \citet{Amaya2011} & 2011 & Memetic cooperative & \cite{Amaya2011} \\ 
 \citet{Amaya2012} & 2012 & Memetic algorithms & \cite{Amaya2012} \\
 \citet{Amaya2013} & 2013 & Cross entropy-based memetic algorithms & \cite{Amaya2013} \\ 
 \citet{Burger2015} & 2015 & Heuristics & \cite{Burger2015} \& Real life \\
 \citet{Catanzaro2015} & 2015 & Exact methods & \cite{Catanzaro2015} \\
 \citet{Chaves2016} & 2016 & Hybrid metaheuristics & \cite{Crama1994, Yanasse2009} \\
 \citet{Paiva2017} & 2017 & Iterated local search & \cite{Crama1994, Catanzaro2015, Yanasse2009} \\ 
 \citet{Ahmadi2018} & 2018 & Hybrid metaheuristics & \cite{Crama1994, Catanzaro2015} \\ 
 \hline
\end{tabular}
\label{tab:works}
\end{table}

Most of the aforementioned GA-based implementations use traditional parent-selection strategies \cite{Whitley2019} such as roulette wheel \cite{Ahmadi2018} and binary tournament \cite{Amaya2008, Amaya2011, Amaya2012}. Survivor selections are solely based on individual objective values, typically obtained with KTNS \cite{Ahmadi2018, Amaya2008, Amaya2011, Amaya2013, Amaya2012, Chaves2016}. \citet{Ahmadi2018} also take into account a similarity function for chromosomes to remove individuals from the population. The crossover operators applied are parameterized uniform crossover \cite{Chaves2016}, uniform cycle crossover \cite{Amaya2012}, alternating position (APX) \cite{Amaya2008, Amaya2011} and partially mapped (PMX) \cite{Ahmadi2018}. Finally, although some authors have developed local search procedures based on problem-specific metrics such as 1-blocks and 0-blocks \citep{Crama1994}, ties arising during move evaluations are generally handled arbitrarily \citep{Amaya2011, Amaya2012}.

We refer the reader to \citet{Camels2019} for an extensive literature review and classification of SSP variants. To date, the best metaheuristic algorithms are those of \citet{Paiva2017}, \citet{Chaves2016} and \citet{Ahmadi2018}. During our experimental comparisons, we will therefore compare our results with the results reported by these authors.

\section{Proposed Methodology}
\label{sec:proposedmethodology}

To effectively solve the SSP, we exploit the hybrid generic search (HGS) paradigm \citep{Vidal2012, Vidal2014} outlined in Algorithm~\ref{alg:hgsssp}. Our algorithm starts with a population of size $\mu$ containing random solutions (permutations) improved by local search (line \ref{lst:line:popinit}). Then, iteratively, it selects two parents by \textit{binary tournament} (line \ref{lst:line:bintourn}) and crosses them via an order crossover (OX) \cite{Oliver1987} to produce a single offspring (line \ref{lst:line:crossover}), which is improved by local search (line \ref{lst:line:ls}). The generation scheme is pursued until the population reaches a maximum size of $|\mathcal{P}| = \mu + \lambda$ individuals. At this point, a survivor selection procedure is applied to discard $\lambda$ individuals. To that extent, the method considers not only the objective value of each individual, but also its contribution to the diversity of the population. The method terminates as soon as $I_{MAX}$ consecutive iterations (individual generations) have been made without improvement of the best solution.

\begin{algorithm}
\caption{HGS with diversity management for the SSP}
\begin{algorithmic}[1]
\STATE Generate an initial population $\mathcal{P}$ with $\mu$ random individuals subject to local search \label{lst:line:popinit}
\WHILE{termination criterion is not attained} \label{lst:line:iter}
\label{lst:line:iterations}
{
    \STATE Select two parents $S_1$ and $S_2$ by binary tournament \label{lst:line:bintourn}
    \STATE Generate a single child $S$ by order crossover (OX) on $S_1$ and $S_2$ \label{lst:line:crossover}
    \STATE Apply local search on $S$ \label{lst:line:ls}
    \STATE Insert the resulting individual into the population $\mathcal{P}$

    \STATE \textbf{if} $|\mathcal{P}| = \mu + \lambda$ \textbf{then} Use a survivors selection procedure to discard $\lambda$ individuals, taking into account their quality and contribution to the population diversity
 
}
\ENDWHILE
\RETURN best solution found
\label{lst:line:return}
\end{algorithmic}
\label{alg:hgsssp}
\end{algorithm}

As visible in Algorithm~\ref{alg:hgsssp}, the overall solution approach follows a simple scheme similar to \cite{Vidal2012, Vidal2014}. Nevertheless, each component and operator (solution evaluation, crossover, local search, diversity management) has been tailored to the SSP and plays a major role in the success of the method. These components will be detailed in the following sections.

\subsection{Solution evaluation}
\label{sec:soleval}

Each solution (i.e., individual) in the population and local search is represented as a simple permutation of jobs.
To evaluate it, we apply the KTNS algorithm \citep{TangAndDenardo1988} as a solution decoder to find an optimal tooling strategy. As illustrated in \autoref{tab:instance} on an example containing $n=10$ jobs, $m=10$ tools with a magazine of capacity $C=4$, the job sequence can be represented as a binary matrix $M$ in which columns are the jobs and rows are the tools. If the $j^\text{th}$ job in the sequence requires tool $t$, then $M_{tj} = 1$, otherwise $M_{tj} = 0$. Since the number of tools needed by each job cannot exceed the magazine capacity, each column contains no more than $C$ times the value $1$.

\begin{table}[htbp]
\centering
\caption{Required tools matrix -- Before KTNS}
\footnotesize{}
\begin{tabular}{ c| c| c c c c c c c c c c }
\hline
 \multicolumn{1}{c}{} & \multicolumn{11}{c}{Jobs}\\
 \hline
\multirow{11}{*}{\rotatebox[origin=c]{90}{Tools}}
 & & 1 & 2 & 3 & 4 & 5 & 6 & 7 & 8 & 9 & 10 \\
 \hline
    & 1 & 0 & 1 & 0 & 0 & 0 & 0 & 0 & 0 & 0 & 0 \\
    & 2 & 1 & 0 & 0 & 0 & 1 & 1 & 1 & 0 & 0 & 0 \\
    & 3 & 0 & 1 & 0 & 0 & 1 & 1 & 0 & 0 & 0 & 0 \\
    & 4 & 0 & 0 & 1 & 1 & 0 & 0 & 1 & 0 & 1 & 0 \\
    & 5 & 0 & 0 & 0 & 0 & 1 & 1 & 0 & 0 & 0 & 0 \\
    & 6 & 1 & 0 & 0 & 0 & 0 & 0 & 0 & 1 & 0 & 0 \\
    & 7 & 0 & 0 & 1 & 0 & 1 & 0 & 0 & 0 & 0 & 1 \\
    & 8 & 0 & 0 & 0 & 1 & 0 & 0 & 0 & 1 & 0 & 1 \\
    & 9 & 0 & 1 & 0 & 0 & 0 & 1 & 1 & 1 & 1 & 0 \\
    & 10 & 0 & 0 & 0 & 0 & 0 & 0 & 0 & 0 & 1 & 0 \\
 \hline
\end{tabular}
\label{tab:instance}
\end{table}

\noindent
\textbf{Evaluation of the primary objective.}
The KTNS algorithm determines the extra tools which may be maintained in the magazine when performing each job. It iterates over the job sequence from beginning to end and adheres to two simple rules:
\begin{enumerate}[nosep]
\item[(i)] at each step, only the tools required for the current job are inserted; and 
\item[(ii)] whenever new tools are loaded, the other tools kept in the machine are those which are needed soonest.
\end{enumerate}
To efficiently implement this policy, we maintain for each job an auxiliary data structure which keeps track, at each instant, of the next instant in which this job is needed. \autoref{tab:loadedmatrix} displays the loaded tools matrix obtained with KTNS on the previous example. For each job, the tools which are kept but not required are represented by an underscored~1.

\begin{table}[htbp]
\centering
\caption{Loaded tools matrix -- After KTNS}
\footnotesize{}
\begin{tabular}{ c| c| c c c c c c c c c c }
\hline
 \multicolumn{1}{c}{} & \multicolumn{11}{c}{Jobs}\\
 \hline
    \multirow{11}{*}{\rotatebox[origin=c]{90}{Tools}}
    & & 1 & 2 & 3 & 4 & 5 & 6 & 7 & 8 & 9 & 10 \\
 \hline
    & 1 & 0 & 1 & 0 & 0 & 0 & 0 & 0 & 0 & 0 & 0 \\
    & 2 & 1 & \underline{1} & \underline{1} & 0 & 1 & 1 & 1 & 0 & 0 & 0 \\
    & 3 & 0 & 1 & \underline{1} & \underline{1} & 1 & 1 & 0 & 0 & 0 & 0 \\
    & 4 & 0 & 0 & 1 & 1 & 0 & 0 & 1 & \underline{1} & 1 & 0 \\
    & 5 & 0 & 0 & 0 & 0 & 1 & 1 & \underline{1} & 0 & 0 & 0 \\
    & 6 & 1 & 0 & 0 & 0 & 0 & 0 & 0 & 1 & 0 & 0 \\
    & 7 & 0 & 0 & 1 & \underline{1} & 1 & 0 & 0 & 0 & 0 & 1 \\
    & 8 & 0 & 0 & 0 & 1 & 0 & 0 & 0 & 1 & \underline{1} & 1 \\
    & 9 & 0 & 1 & 0 & 0 & 0 & 1 & 1 & 1 & 1 & \underline{1} \\
    & 10 & 0 & 0 & 0 & 0 & 0 & 0 & 0 & 0 & 1 & \underline{1} \\
 \hline
\end{tabular}
\label{tab:loadedmatrix}
\end{table}

The number of tool switches can be obtained from \autoref{tab:loadedmatrix} by computing the number of times a $1$ turns into a $0$ in every row, which is equivalent to replacing a tool by another in the machine.\\

\noindent
\textbf{Evaluation of the tie-breaking objective.}
Many solutions explored during the search, within the genetic algorithm and the local search, have the same number of tool switches. It is therefore necessary to guide the algorithm in ``plateau'' regions of the search space, where all solutions have the same primary objective. To that extent, we define a secondary objective, designed to break ties in favor of solutions which are more likely to lead to future improvements. We use the concept of 0-blocks, first introduced in \citet{Crama1994}. A 0-block is a maximum sequence of consecutive zeros, preceded and followed by a one, in the loaded matrix representation. It represents an interval during which a tool is not needed, but which could be filled by maintaining the tool in the magazine. In \autoref{tab:loadedmatrix}, for example, the jobs in positions 5 to 7 represent a 0-block of size 3 for tool 8. Intuitively, short 0-blocks are more likely to be filled during the search process, leading to an effective reduction of the number of tool switches and to an improvement of the primary objective.
For a solution~$S$, and for each tool $j$, let $k_j(S)$ be the number of 0-blocks in the loaded tool matrix and let $b^x_j(S)$ be the size of the $x^\text{th}$ 0-block, for $x \in \{1,\dots,k_j(S)\}$. The tie-breaking objective is then evaluated as:
\begin{equation}
\Phi'(S) = \sum_{j=1}^m \sum_{x=1}^{k_j(S)} \sqrt{b^x_j(S)}.
\end{equation}
We use the square root function due to its concavity. By minimizing $\Phi'(S)$, we favor solutions with short and large 0-blocks (e.g., one block of size 2 and another of size 8) over solutions with balanced blocks (e.g., two blocks of size 5), with the aim of ultimately eliminating some of the shortest ones and reducing the number of tool switches.

\subsection{Generation of new solutions}

Firstly, two parents are selected by binary tournament. Each binary tournament selection consists in pickup randomly (with uniform probability) two individuals in the population and retaining the one with the best biased fitness, as defined in Section \ref{sec:bf}.

\begin{figure}[!htbp]
    \centering
    \includegraphics[scale=0.75]{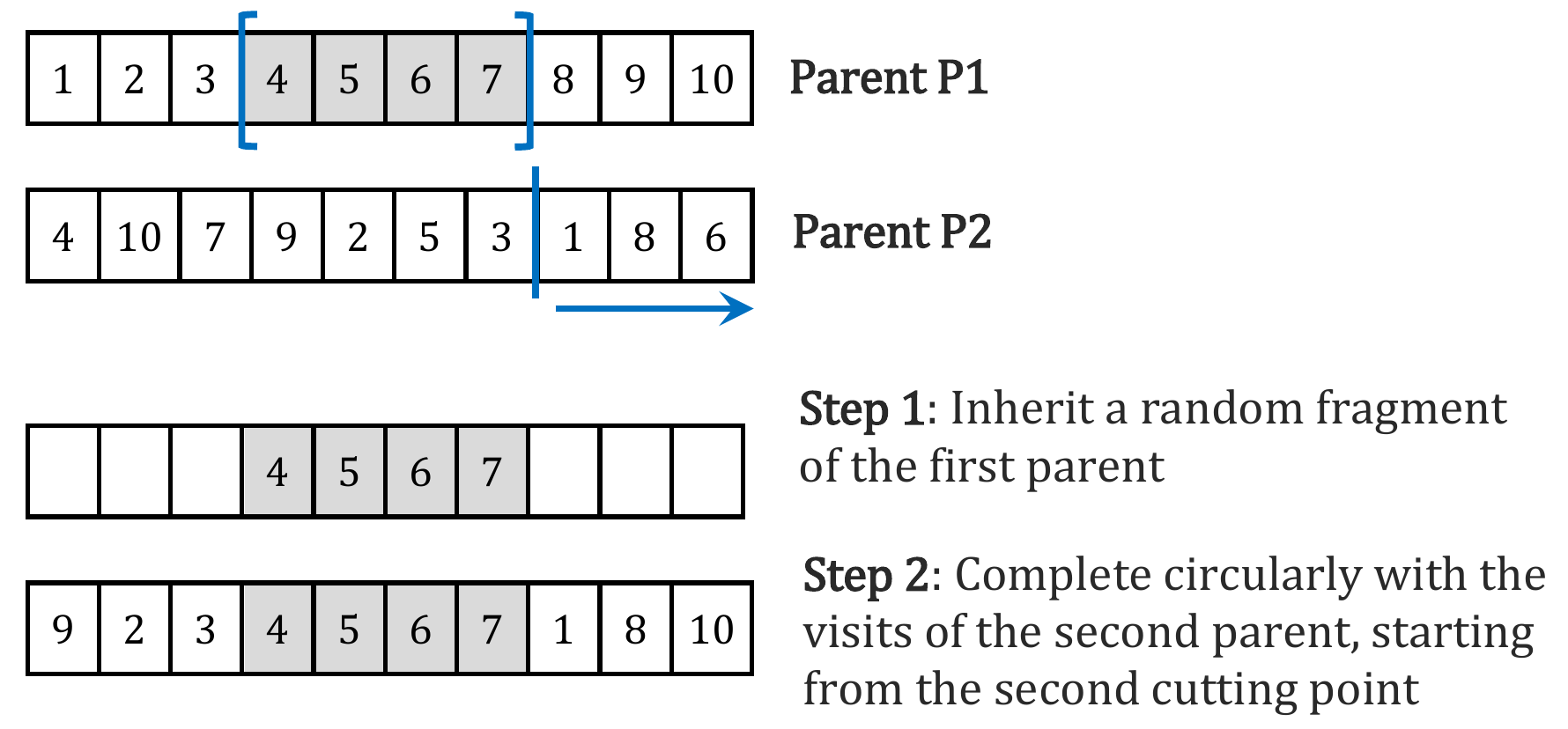}
    \caption{Illustration of the OX crossover on a small example with $10$ jobs}
    \label{figOX}
\end{figure}

Secondly, the order crossover (OX) \citep{Oliver1987} is applied on the two parents to generate a single child. This crossover, illustrated in Figure \ref{figOX}, consists in (i) selecting a random substring from the first parent ; (ii) copying this substring into the child while leaving the rest of the positions empty; and (iii) sweeping through the second parent to fill the empty positions with the missing visits.

Finally, the resulting offspring is improved by means of three successive local searches, based on the \textsc{2-opt}, \textsc{relocate} and \textsc{swap} neighborhoods, in this specific order. As discussed in Section \ref{sec-sensitivity}, we performed sensitivity analysis for each alternative neighborhood ordering, and this setting led to solutions of significantly higher quality. In each local search, the neighborhood is explored in random order according to a first-improvement policy, i.e., any improving move (improving the number of tool switches, or improving the tie-breaking objective for an equal number of tool switches) is immediately applied. Each local search stops as soon as all moves have been successively tested without improvement of the primary or auxiliary objective, therefore reaching a local minimum. The resulting solution is added into the population.

\subsection{Biased fitness and population diversity management}
\label{sec:bf}

The biased fitness $f_{\mathcal{P}}(p)$ of each individual $p$ in the population $ \mathcal{P}$ is calculated in a similar way as in \cite{Vidal2012, Vidal2014}. The individuals are kept ordered in terms of quality (considering primary and tie-breaking objectives) and diversity contribution. To calculate the diversity contribution of each individual, the algorithm computes its average distance to its $\mu^{close}$ closest ones, the distance between two individuals (represented as job permutations) being defined as the number of different edges between them (broken-pairs distance). Then, it sorts the population in decreasing order of diversity contribution, and in increasing order of objective function, associating to each individual a diversity rank $f_{\mathcal{P}}^{div}(p)$ and a quality rank $f_{\mathcal{P}}^{obj}(p)$. Finally, the biased fitness of each individual is calculated as:
\begin{equation}
f_{\mathcal{P}}(p) = f_{\mathcal{P}}^{obj}(p) + \left( 1 - \frac{\mu^{elite}}{\mathcal{\mid P \mid}} \right) \times f_{\mathcal{P}}^{div}(p)
\end{equation}
Small values of biased fitness correspond to promising individuals, with a small objective value and a large contribution to the population diversity.

The biased fitness measure is used when selecting parents by binary tournament (retaining the individual with the smallest biased fitness) and when selecting individuals to exclude from the population during survivor selections. In the latter case, the algorithm iteratively excludes the worst individual (i.e., with highest biased fitness) having a clone whenever duplicated solutions exist in the population, or the worst individual otherwise. This process is repeated until the desired population size of $\mu$ is attained. As discussed in \cite{Vidal2012}, this survivor selection procedure preserves diversity and meanwhile guarantees that the best $\mu^{elite}$ individuals in the population remain preserved.

\section{Computational Experiments}
\label{sec:computationalexperiments}

We implemented the proposed method in C++ and conducted our experiments on a single thread of an Intel Core i5-4288U 2.6\,GHz processor with 8\,GB of RAM running macOS High Sierra 10.13.6. It was compiled with g++ 8.1 using the -O3 flag. The source code and benchmark instances are also accessible at \url{https://github.com/jordanamecler/HGS-SSP}. {\color{blue}[Password protected until publication]}.

We first describe the characteristics of the instances used in our experiments. Next, we explain how we calibrated the parameters of the algorithm and investigate the impact of several methodological choices. Finally, we compare our algorithm with the current state-of-the-art methods and provide solutions for a new set of larger instances.

\subsection{Benchmark instances}

As indicated in \autoref{tab:ssp_instances}, we use the classical benchmark instances of \citet{Crama1994}, \citet{Yanasse2009}, and \citet{Catanzaro2015} to evaluate the performance of the proposed method. These instances are organized in different groups corresponding to different size parameters. Groups $A$, $B$, $C$, $D$, and $E$ (from \cite{Yanasse2009}) contain a total of 1350 instances with 8 to 25 jobs. Groups $C_1, C_2, C_3$, and $C_4$ (from \cite{Crama1994}) as well as $datA$, $datB$, $datC$, and $datD$ (from \cite{Catanzaro2015}) contain 40 instances each and include 10 to 40 jobs.
    
    \begin{table}[!htbp]
    \centering
\renewcommand{\arraystretch}{1.2}    
\caption{SSP instances}
    \footnotesize{}
    \begin{tabular}{ l l l l l l l l l l }
     \hline
     \multirow{2}{*}{\textbf{Group}} & \multicolumn{2}{c}{\textbf{\#Jobs}} & & \multicolumn{2}{c}{\textbf{\#Tools}} & & \multicolumn{2}{c}{\textbf{Capacity}} & \multirow{2}{*}{\textbf{\#Instances}} \\
     \cline{2-3}
     \cline{5-6}
     \cline{8-9}
      & \textbf{Min} & \textbf{Max} & & \textbf{Min} & \textbf{Max} & & \textbf{Min} & \textbf{Max} & \\
     \hline
     $A$ & 8 & 8 & & 15 & 25 & & 5 & 20 & 340 \\
     $B$ & 9 & 9 & & 15 & 25 & & 5 & 20 & 330 \\
     $C$ & 15 & 15 & & 15 & 25 & & 5 & 20 & 340 \\
     $D$ & 20 & 25 & & 15 & 25 & & 5 & 20 & 260 \\
     $E$ & 10 & 15 & & 10 & 20 & & 4 & 12 & 80 \\
     $C_1$ & 10 & 10 & & 10 & 10 & & 4 & 7 & 40 \\
     $C_2$ & 15 & 15 & & 20 & 20 & & 6 & 12 & 40 \\
     $C_3$ & 30 & 30 & & 40 & 40 & & 15 & 25 & 40 \\
     $C_4$ & 40 & 40 & & 60 & 60 & & 20 & 30 & 40 \\
     $datA$ & 10 & 10 & & 10 & 10 & & 4 & 7 & 40 \\
     $datB$ & 15 & 15 & & 20 & 20 & & 6 & 12 & 40 \\
     $datC$ & 30 & 30 & & 40 & 40 & & 15 & 25 & 40 \\
     $datD$ & 40 & 40 & & 60 & 60 & & 20 & 30 & 40 \\
     \hline
    \end{tabular}
    \label{tab:ssp_instances}
    \end{table}

\subsection{Parameters Calibration}

We first conducted preliminary experiments on a subset of instances to find suitable parameter values for our algorithm. We chose to perform the calibration tests on the benchmark of \citet{Crama1994} since it contains instances with very diverse characteristics. This led to our \emph{baseline configuration} with the following parameter values: $\mu = 20$, $\lambda = 40$, $\mu^{elite} = 10$, and $\mu^{close} = 3$. Subsequently, we performed additional analyses to investigate the impact of any deviation from this parameter setting. We modify the value of each parameter using a one-factor-at-a-time (OFAT) approach and report the results obtained by each configuration.

Tables \ref{tab:pop_size}, \ref{tab:elite}, and \ref{tab:close_individuals} show the average results of different method configurations for the benchmark instances of \citet{Crama1994}. In these tables, \textbf{Avg} is the average objective value, \textbf{T} is the average CPU time in seconds, and \textbf{Gap (\%)} is the percentage gap, calculated as $100 \times \big(\frac{\text{Avg} - \text{BKS}}{\text{BKS}}\big)$, where BKS is the best known solution collected from all previous studies. As visible in these experiments, modifying $\mu^{close}$ has only a minor impact on the solution quality and CPU time. On the other hand, as $\mu^{elite}$ increases, the CPU time tends to increase, and the best results are obtained when it is set to half of the base population size. The CPU time increases with the population size, but a mid-sized population yielded the best results in terms of solution quality.

\begin{table}[htbp]
\centering
\renewcommand{\arraystretch}{1.2}
\caption{Varying population size}
\footnotesize{}
\begin{tabular}{ l l l l l l } 
 \hline
 $\mu$ & $\lambda$ & $\mu^{elite}$ & $\mu^{close}$ &  \multicolumn{1}{c}{Avg}  & \multicolumn{1}{c}{T} \\
 \hline
10 & 10 & 5 & 2 & 53.07 & 105.35 \\
10 & 20 & 5 & 2 & 53.04 & 112.70 \\
20 & 20 & 10 & 3 & 53.01 & 127.12 \\
20 & 40 & 10 & 3 & \bftab{53.00} & 133.09 \\
40 & 40 & 20 & 5 & 53.00 & 152.58 \\
40 & 80 & 20 & 5 & 53.02 & 163.27 \\
80 & 80 & 40 & 10 & 53.04 & 181.49 \\
80 & 160 & 40 & 10 & 53.06 & 190.17 \\
 \hline
\end{tabular}
\label{tab:pop_size}
\end{table}

\begin{table}[htbp]
\centering
\renewcommand{\arraystretch}{1.2}
\caption{Varying $\mu^{elite}$}
\footnotesize{}
\begin{tabular}{ l l l l l l } 
 \hline
 $\mu$ & $\lambda$ & $\mu^{elite}$ & $\mu^{close}$ & \multicolumn{1}{c}{Avg}  & \multicolumn{1}{c}{T} \\
 \hline
20 & 40 & 2 & 3 & 53.03 & 160.65 \\
20 & 40 & 5 & 3 & 53.02 & 144.80 \\
20 & 40 & 8 & 3 & 53.01 & 139.42 \\
20 & 40 & 10 & 3 & \bftab{53.00} & 133.09 \\
20 & 40 & 12 & 3 & 53.00 & 129.29 \\
20 & 40 & 15 & 3 & 53.02 & 124.60 \\
20 & 40 & 20 & 3 & 53.03 & 114.59 \\
 \hline
\end{tabular}
\label{tab:elite}
\end{table}

\begin{table}[htbp]
\centering
\renewcommand{\arraystretch}{1.2}
\caption{Varying $\mu^{close}$}
\footnotesize{}
\begin{tabular}{ l l l l l l }
 \hline
 $\mu$ & $\lambda$ & $\mu^{elite}$ & $\mu^{close}$ & \multicolumn{1}{c}{Avg}  & \multicolumn{1}{c}{T} \\
 \hline
20 & 40 & 10 & 1 & 53.02 & 132.28 \\
20 & 40 & 10 & 2 & 53.00 & 133.16 \\
20 & 40 & 10 & 3 & \bftab{53.00} & 133.09 \\
20 & 40 & 10 & 5 & 53.00 & 138.38 \\
20 & 40 & 10 & 7 & 53.00 & 138.33 \\
20 & 40 & 10 & 10 & \bftab{53.00} & 138.22 \\
 \hline
\end{tabular}
\label{tab:close_individuals}
\end{table}

\subsection{Sensitivity analysis}
\label{sec-sensitivity}

We performed a sensitivity analysis for the main components of the algorithm. The following modifications of the method were tested: changing the order of the neighborhoods, deactivating some or several neighborhoods, deactivating the diversity management component, and deactivating the secondary objective. The results of these alternative configurations are reported in \autoref{tab:sensitivityanalysis} and compared to our baseline configuration.

\begin{table}[htbp!]
\centering
\renewcommand{\arraystretch}{1.2}
\caption{Sensitivity Analysis}
\footnotesize{}
\begin{tabular}{ l l l l l }
 \hline
 Configuration & Avg & Gap(\%) & T \\
 \hline
Baseline & 53.00 & \bftab{-0.01} & 135.31\\
\textsc{2opt}-\textsc{swap}-\textsc{relocate} & 53.01 & 0.02 & 145.42 \\
\textsc{swap}-\textsc{2opt}-\textsc{relocate} & 53.06 & 0.09 & 178.56 \\
\textsc{swap}-\textsc{relocate}-\textsc{2opt} & 53.06 & 0.10 & 183.41 \\
\textsc{relocate}-\textsc{2opt}-\textsc{swap} & 53.06 & 0.09 & 160.73 \\
\textsc{relocate}-\textsc{swap}-\textsc{2opt} & 53.08 & 0.12 & 183.61 \\
without \textsc{2opt} & 53.61 & 0.67 & 130.95 \\
without \textsc{swap} & 53.02 & 0.02 & 118.97 \\
without \textsc{relocate} & 53.11 & 0.14 & 105.70 \\
without \textsc{2opt} and \textsc{swap} & 53.80 & 0.88 & 74.93 \\
without \textsc{2opt} and \textsc{relocate} & 53.66 & 0.72 & 71.03 \\
without \textsc{relocate} and \textsc{swap} & 53.17 & 0.20 & 74.02 \\
without diversity management & 53.04 & 0.18 & 111.63 \\
without secondary objective & 53.16 & 0.22 & 87.20 \\
 \hline
\end{tabular}
\label{tab:sensitivityanalysis}
\end{table}

These results confirm our methodological design and baseline parameter choices. Indeed, all these alternative configurations obtained solutions of significantly worse quality for only moderate time gains. The \textsc{2opt} neighborhood appears to be the most important for a good performance, followed by the \textsc{relocate} and \textsc{swap} neighborhoods. This difference of impact may also explain why the exploration of the neighborhoods by order of importance \textsc{2opt}-\textsc{relocate}-\textsc{swap} (as done in our baseline configuration) leads to generally better results in shorter time. The use of the secondary objective contributes to attain solutions of significantly higher quality, and the diversity management strategy has a smaller but still beneficial impact on solution quality.

\begin{figure}[htbp]
  \vspace*{0.8cm}
  \hspace*{-0.65cm}
  \scalebox{1.2}
  {
  \begin{minipage}[b]{0.43\textwidth}
    \includegraphics[width=\textwidth]{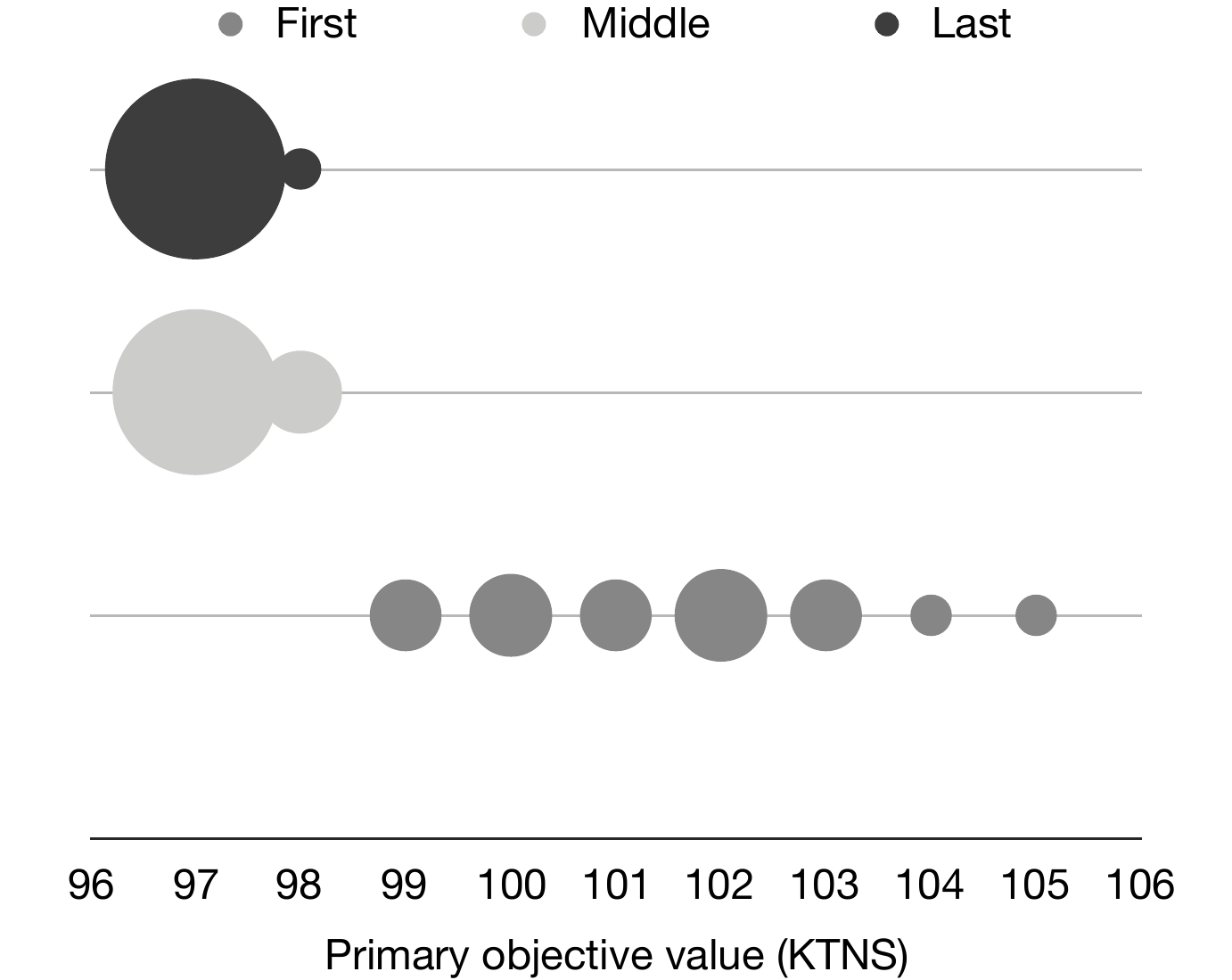}
  \end{minipage}
  \hfill
  \begin{minipage}[b]{0.43\textwidth}
    \includegraphics[width=\textwidth]{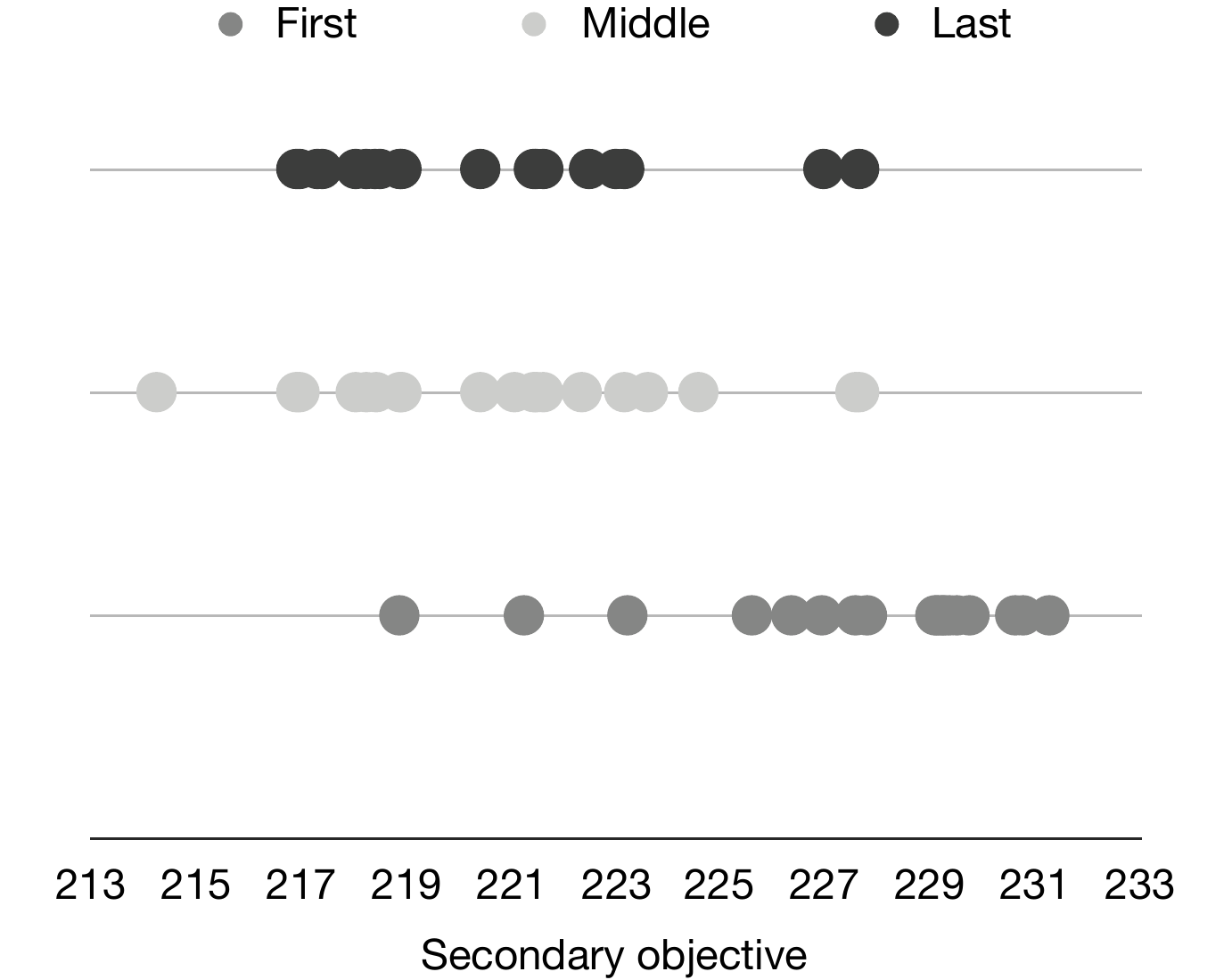}
  \end{minipage}
  }
  \caption{Distribution of objective values in the population, at different stages of the search}
  \label{fig:dist}
\end{figure}

To better visualize the impact of the secondary objective, we conducted another experimental analysis of the distribution of the individuals in the objective space during a typical GA run, on the first instance of group $C_3$. This analysis is displayed in Figure \ref{fig:dist} at three stages of the solution process: after the first survivors selection, half way through the execution, and after the last survivors selection. On this figure, dot sizes are proportional to the number of solutions sharing the same objective value. Visibly, numerous solutions share the same primary (KTNS) objective value, especially at intermediate or late stages of the search. Fortunately, secondary objective values are better spread, therefore allowing to rank solutions and guide the search towards more promising regions by selection pressure.

\subsection{Comparison with other algorithms}

The tables presented hereafter report the best solution (\textbf{Best}), the average solution over 10 runs (\textbf{Avg}), and the average CPU time in seconds (\textbf{T}) of each method. The best solutions, for each group of instances, are highlighted in boldface.
It is worth mentioning that new random seeds were used in these experiments to eliminate any possibility of overtuning.\\

\noindent
\textbf{Comparison with previous heuristics.}
We compare our results with those found by the best methods from the literature, namely the CS+BRKGA of \citet{Chaves2016}, the ILS of \citet{Paiva2017} and the DQGA of \citet{Ahmadi2018}. The following experimental setup has been used in previous studies, based on the data reported in previous papers and additional information sent to us by the authors:
\begin{itemize}[nosep]
    \item \citet{Chaves2016}: average of 20 runs on an Intel i7-2600 3.4\,GHz  with 16\,GB of RAM.
    \item \citet{Paiva2017}: average of 20 runs on an Intel i5-3330 3.2\,GHz with 8\,GB~of~RAM. 
    \item \citet{Ahmadi2018}: average of 10 runs on an Intel i7 3.4\,GHz  with 16\,GB of RAM.
\end{itemize}
For a fair comparison with these studies, we converted all CPU times into their equivalent time on our i5-4288U 2.6\,GHz processor. To that extent, we used the single thread rating from Passmark benchmark (\url{www.cpubenchmark.net}), leading to conversion factors of 1.0977, 0.9709 and 1.1394 for the machines used in \citet{Chaves2016}, \citet{Paiva2017} and \citet{Ahmadi2018}, respectively. Moreover, since the precise processor model used in \citet{Ahmadi2018} is not documented, we used the average rating of two popular Intel i7 3.4\,GHz models of the same generation as a good approximation. 

Table \ref{tab:summary} presents a summary of the results in which the results are aggregated per instance group. The proposed HGS-SSP found the solutions of best quality for all groups, in a computational time which is equal of smaller than existing algorithms. The difference of solution quality and speed is especially visible on the larger and more challenging instances of groups $C_3$, $C_4$, $datC$ and $datD$.

\begin{table}[htbp!]
\centering
\renewcommand{\arraystretch}{1.2}
\setlength{\tabcolsep}{2.5pt}
\caption{Summary of results}
\footnotesize{}
\begin{tabular}{ l l l l l l l l l l l l l l l l }
\hline
& \multicolumn{3}{c}{CS+BRKGA} & & \multicolumn{3}{c}{ILS} & & \multicolumn{3}{c}{DQGA} & & \multicolumn{3}{c}{HGS-SSP}\\
\cline{2-4}
\cline{6-8}
\cline{10-12}
\cline{14-16}
Group & Best & Avg & T & & Best & Avg & T & & Best & Avg & T & & Best & Avg & T \\
\hline
$A$ & \bftab{12.63} & \bftab{12.63} & 4.07 & & \bftab{12.63} & \bftab{12.63} & 0.09 & & - & - & - & & \bftab{12.63} & \bftab{12.63} & 0.06 \\
$B$ & \bftab{13.32} & 13.32 & 4.45 & & \bftab{13.32} & 13.32 & 0.15 & & - & - & - & & \bftab{13.32} & \bftab{13.32} & 0.10 \\
$C$ & \bftab{17.46} & 17.62 & 10.80 & & \bftab{17.46} & 17.46 & 1.40 & & - & - & - &  & \bftab{17.46} & 17.46 & 0.91 \\
$D$ & 13.40 & 13.68 & 30.37 & & \bftab{13.38} & 13.39 & 5.07 & & - & - & - & &  \bftab{13.38} & 13.38 & 4.28 \\
$E$ & \bftab{9.64} & 9.70 & 7.18 & & \bftab{9.64} & \bftab{9.64} & 0.39 & & - & - & - & & \bftab{9.64} & 9.64 & 0.43 \\
$C_1$ & \bftab{5.68} & 5.68 & 2.42 & & \bftab{5.68} & \bftab{5.68} & 0.07 & & \bftab{5.68} &\bftab{5.68} & 0.26 & & \bftab{5.68} & \bftab{5.68} & 0.07 \\
$C_2$ & \bftab{13.00} & 13.07 & 11.58 & & \bftab{13.00} & 13.01 & 0.83 & & \bftab{13.00} & 13.03 & 1.38 & & \bftab{13.00} & \bftab{13.00} & 0.78 \\ 
$C_3$ & 60.58 & 61.71 & 123.15 & & 60.25 & 60.54 & 130.74 & & 60.15 & 60.99 & 213.22 & & \bftab{60.08} & 60.17 & 67.45 \\
$C_4$ & 135.03 & 137.21 & 541.10 & & 133.70 & 134.19 & 874.98 & & 133.63 & 136.28 & 940.01 & & \bftab{132.78} & 132.99 & 472.55 \\
$datA$ & - & - & - & & \bftab{5.35} & \bftab{5.35} & 0.07 & & \bftab{5.35} &\bftab{5.35} & 0.21 & & \bftab{5.35} & \bftab{5.35} & 0.06 \\
$datB$ & - & - & - & & \bftab{12.78} & \bftab{12.78} & 1.05 & & \bftab{12.78} & 12.86 & 1.69 & & \bftab{12.78} & \bftab{12.78} & 0.74 \\ 
$datC$ & - & - & - & & 55.53 & 55.79 & 132.54 & & 55.53 & 56.62 & 201.21 & & \bftab{55.43} & 55.48 & 63.77 \\ 
$datD$ & - & - & - & & 134.00 & 134.38 & 917.35 & & 133.98 & 136.19 & 736.50 & & \bftab{132.85} & 133.11 & 472.33 \\
 \hline
\end{tabular}
\label{tab:summary}
\end{table}

Tables \ref{tab:groupA} to \ref{tab:groupdat} now compare the results of our HGS with those of the best methods from the literature for each particular group.
We therefore compare with CS+BRKGA and ILS for the first five groups (A, B, C, D and E), and with  DQGA and ILS for the remaining groups.

\begin{table}[htbp!]
\centering
\renewcommand{\arraystretch}{1.2}
\setlength{\tabcolsep}{4pt}
\caption{Group $A$}
\footnotesize{}
\begin{tabular}{ l l l l l l l l l l l l l l l }
\hline
& & & & \multicolumn{3}{c}{CS+BRKGA} & & \multicolumn{3}{c}{ILS} & & \multicolumn{3}{c}{HGS-SSP}\\
\cline{5-7}
\cline{9-11}
\cline{13-15}
$n$ & $m$ & $C$ & $i$ & Best & Avg & T & & Best & Avg & T & & Best & Avg & T \\
\hline
8 & 15 & 5 & 10 & \bftab{12.00} & \bftab{12.00} & 2.62 & & \bftab{12.00} & \bftab{12.00} & 0.04 & & \bftab{12.00} & \bftab{12.00} & 0.04 \\
8 & 15 & 10 & 30 & \bftab{6.83} & \bftab{6.83} & 2.91 & & \bftab{6.83} & \bftab{6.83} & 0.07 & & \bftab{6.83} & \bftab{6.83} & 0.04 \\
8 & 20 & 5 & 10 & \bftab{16.80} & \bftab{16.80} & 3.26 & & \bftab{16.80} & \bftab{16.80} & 0.06 & & \bftab{16.80} & \bftab{16.80} & 0.06 \\
8 & 20 & 10 & 30 & \bftab{13.07} & \bftab{13.07} & 3.89 & & \bftab{13.07} & \bftab{13.07} & 0.13 & & \bftab{13.07} & \bftab{13.07} & 0.06 \\
8 & 20 & 15 & 60 & \bftab{7.08} & \bftab{7.08} & 3.92 & & \bftab{7.08} & \bftab{7.08} & 0.10 & & \bftab{7.08} & \bftab{7.08} & 0.05 \\
8 & 25 & 5 & 10 & \bftab{20.10} & \bftab{20.10} & 4.99 & & \bftab{20.10} & \bftab{20.10} & 0.03 & & \bftab{20.10} & \bftab{20.10} & 0.07 \\
8 & 25 & 10 & 30 & \bftab{18.20} & \bftab{18.20} & 4.80 & & \bftab{18.20} & \bftab{18.20} & 0.12 & & \bftab{18.20} & \bftab{18.20} & 0.08 \\
8 & 25 & 15 & 60 & \bftab{12.95} & \bftab{12.95} & 5.06 & & \bftab{12.95} & \bftab{12.95} & 0.14 & & \bftab{12.95} & \bftab{12.95} & 0.08 \\
8 & 25 & 20 & 100 & \bftab{6.61} & \bftab{6.61} & 5.18 & & \bftab{6.61} & \bftab{6.61} & 0.11 & & \bftab{6.61} & \bftab{6.61} & 0.06 \\
 \hline
\end{tabular}
\label{tab:groupA}
\end{table}

\begin{table}[htbp!]
\centering
\renewcommand{\arraystretch}{1.2}
\setlength{\tabcolsep}{4pt}
\caption{Group $B$}
\footnotesize{}
\begin{tabular}{ l l l l l l l l l l l l l l l }
\hline
& & & & \multicolumn{3}{c}{CS+BRKGA} & & \multicolumn{3}{c}{ILS} & & \multicolumn{3}{c}{HGS-SSP}\\
\cline{5-7}
\cline{9-11}
\cline{13-15}
$n$ & $m$ & $C$ & $i$ & Best & Avg & T & & Best & Avg & T & & Best & Avg & T \\
\hline
9 & 15 & 5 & 10 & \bftab{12.20} & \bftab{12.20} & 2.99 & & \bftab{12.20} & \bftab{12.20} & 0.07 & & \bftab{12.20} & \bftab{12.20} & 0.07 \\
9 & 15 & 10 & 30 & \bftab{7.37} & \bftab{7.37} & 3.46 & & \bftab{7.37} & \bftab{7.37} & 0.11 & & \bftab{7.37} & \bftab{7.37} & 0.07 \\
9 & 20 & 5 & 10 & \bftab{17.40} & \bftab{17.40} & 3.44 & & \bftab{17.40} & \bftab{17.40} & 0.08 & & \bftab{17.40} & \bftab{17.40} & 0.09 \\
9 & 20 & 10 & 30 & \bftab{14.17} & \bftab{14.17} & 4.30 & & \bftab{14.17} & \bftab{14.17} & 0.18 & & \bftab{14.17} & \bftab{14.17} & 0.10 \\
9 & 20 & 15 & 60 & \bftab{7.60} & \bftab{7.60} & 4.38 & & \bftab{7.60} & \bftab{7.60} & 0.18 & & \bftab{7.60} & \bftab{7.60} & 0.08 \\
9 & 25 & 5 & 10 & \bftab{20.40} & \bftab{20.40} & 4.48 & & \bftab{20.40} & \bftab{20.40} & 0.05 & & \bftab{20.40} & \bftab{20.40} & 0.11 \\
9 & 25 & 10 & 30 & \bftab{18.77} & \bftab{18.77} & 5.58 & & \bftab{18.77} & 18.77 & 0.20 & & \bftab{18.77} & \bftab{18.77} & 0.13 \\
9 & 25 & 15 & 50 & \bftab{14.74} & 14.75 & 5.60 & & \bftab{14.74} & \bftab{14.74} & 0.32 & & \bftab{14.74} & \bftab{14.74} & 0.13 \\
9 & 25 & 20 & 100 & \bftab{7.19} & \bftab{7.19} & 5.78 & & \bftab{7.19} & 7.19 & 0.19 & & \bftab{7.19} & \bftab{7.19} & 0.10 \\
 \hline
\end{tabular}
\label{tab:groupB}
\end{table}

\begin{table}[htbp!]
\centering
\renewcommand{\arraystretch}{1.2}
\setlength{\tabcolsep}{4pt}
\caption{Group $C$}
\footnotesize{}
\begin{tabular}{ l l l l l l l l l l l l l l l }
\hline
& & & & \multicolumn{3}{c}{CS+BRKGA} & & \multicolumn{3}{c}{ILS} & & \multicolumn{3}{c}{HGS-SSP}\\
\cline{5-7}
\cline{9-11}
\cline{13-15}
$n$ & $m$ & $C$ & $i$ & Best & Avg & T & & Best & Avg & T & & Best & Avg & T \\
\hline
15 & 15 & 5 & 10 & \bftab{16.60} & 16.69 & 5.83 & & \bftab{16.60} & \bftab{16.60} & 0.51 & & \bftab{16.60} & \bftab{16.60} & 0.71 \\
15 & 15 & 10 & 30 & \bftab{9.80} & 9.88 & 7.80 & & \bftab{9.80} & \bftab{9.80} & 0.86 & & \bftab{9.80} & \bftab{9.80} & 0.57 \\
15 & 20 & 5 & 10 & \bftab{20.60} & 20.77 & 7.97 & & \bftab{20.60} & \bftab{20.60} & 0.65 & & \bftab{20.60} & \bftab{20.60} & 0.83 \\
15 & 20 & 10 & 30 & \bftab{18.33} & 18.52 & 9.81 & & \bftab{18.33} & \bftab{18.33} & 1.54 & & \bftab{18.33} & \bftab{18.33} & 0.96 \\
15 & 20 & 15 & 60 & \bftab{10.52} & 10.65 & 10.55 & & \bftab{10.52} & 10.52 & 1.66 & & \bftab{10.52} & 10.52 & 0.76 \\
15 & 25 & 5 & 10 & \bftab{27.50} & 27.7 & 10.21 & & \bftab{27.50} & \bftab{27.50} & 0.63 & & \bftab{27.50} & 27.51 & 1.02 \\
15 & 25 & 10 & 30 & \bftab{25.07} & 25.30 & 14.85 & & \bftab{25.07} & 25.07 & 1.86 & & \bftab{25.07} & \bftab{25.07} & 1.28 \\
15 & 25 & 15 & 60 & \bftab{19.07} & 19.27 & 14.97 & & \bftab{19.07} & \bftab{19.07} & 2.93 & & \bftab{19.07} & 19.07 & 1.18 \\
15 & 25 & 20 & 100 & \bftab{9.66} & 9.79 & 15.17 & & \bftab{9.66} & 9.66 & 2.00 & & \bftab{9.66} & 9.66 & 0.90 \\
 \hline
\end{tabular}
\label{tab:groupC}
\end{table}

\begin{table}[htbp!]
\centering
\renewcommand{\arraystretch}{1.2}
\setlength{\tabcolsep}{4pt}
\caption{Group $D$}
\footnotesize{}
\begin{tabular}{ l l l l l l l l l l l l l l l }
\hline
& & & & \multicolumn{3}{c}{CS+BRKGA} & & \multicolumn{3}{c}{ILS} & & \multicolumn{3}{c}{HGS-SSP}\\
\cline{5-7}
\cline{9-11}
\cline{13-15}
$n$ & $m$ & $C$ & $i$ & Best & Avg & T & & Best & Avg & T & & Best & Avg & T \\
\hline
20 & 15 & 5 & 10 & 21.10 & 21.58 & 11.84 & & \bftab{20.90} & \bftab{20.90} & 1.35 & & \bftab{20.90} & \bftab{20.90} & 2.59 \\
20 & 15 & 10 & 20 & \bftab{8.20} & 8.44 & 13.55 & & \bftab{8.20} & 8.21 & 2.13 & & \bftab{8.20} & \bftab{8.20} & 1.83 \\
20 & 20 & 5 & 10 & 24.30 & 24.93 & 16.29 & & \bftab{24.20} & 24.24 & 1.80 & & \bftab{24.20} & \bftab{24.20} & 3.23 \\
20 & 20 & 10 & 10 & \bftab{10.60} & 10.76 & 17.66 & & \bftab{10.60} & \bftab{10.60} & 2.79 & & \bftab{10.60} & \bftab{10.60} & 2.52 \\
20 & 20 & 15 & 30 & \bftab{6.67} & 6.79 & 27.08 & & \bftab{6.67} & \bftab{6.67} & 3.22 & & \bftab{6.67} & \bftab{6.67} & 2.27 \\
20 & 25 & 5 & 10 & \bftab{30.10} & 30.74 & 21.04 & & \bftab{30.10} & \bftab{30.10} & 2.04 & & \bftab{30.10} & 30.11 & 3.99 \\
20 & 25 & 10 & 10 & \bftab{15.40} & 15.47 & 23.60 & & \bftab{15.40} & \bftab{15.40} & 3.62 & & \bftab{15.40} & \bftab{15.40} & 3.37 \\
20 & 25 & 15 & 40 & \bftab{21.25} & 21.75 & 30.87 & & \bftab{21.25} & 21.26 & 7.19 & & \bftab{21.25} & 21.25 & 3.88 \\
20 & 25 & 20 & 40 & \bftab{6.15} & 6.28 & 39.01 & & \bftab{6.15} & \bftab{6.15} & 3.35 & & \bftab{6.15} & \bftab{6.15} & 2.69 \\
25 & 15 & 10 & 10 & \bftab{5.90} & 6.00 & 23.12 & & \bftab{5.90} & \bftab{5.90} & 3.85 & & \bftab{5.90} & \bftab{5.90} & 3.78 \\
25 & 20 & 10 & 10 & \bftab{11.60} & 12.05 & 30.17 & & \bftab{11.60} & 11.61 & 9.44 & & \bftab{11.60} & \bftab{11.60} & 6.58 \\
25 & 20 & 15 & 10 & \bftab{7.60} & 7.82 & 28.18 & & \bftab{7.60} & \bftab{7.60} & 9.63 & & \bftab{7.60} & \bftab{7.60} & 6.74 \\
25 & 25 & 10 & 10 & \bftab{16.60} & 17.06 & 40.49 & & \bftab{16.60} & 16.67 & 11.33 & & \bftab{16.60} & 16.67 & 8.60 \\
25 & 25 & 15 & 10 & \bftab{10.00} & \bftab{10.00} & 60.06 & & \bftab{10.00} & \bftab{10.00} & 7.60 & & \bftab{10.00} & \bftab{10.00} & 6.21 \\
25 & 25 & 20 & 30 & \bftab{5.50} & 5.59 & 72.58 & & \bftab{5.50} & \bftab{5.50} & 6.68 & & \bftab{5.50} & \bftab{5.50} & 5.97 \\
 \hline
\end{tabular}
\label{tab:groupD}
\end{table}

\begin{table}[htbp!]
\centering
\renewcommand{\arraystretch}{1.2}
\setlength{\tabcolsep}{4pt}
\caption{Group $E$}
\footnotesize{}
\begin{tabular}{ l l l l l l l l l l l l l l l }
\hline
& & & & \multicolumn{3}{c}{CS+BRKGA} & & \multicolumn{3}{c}{ILS} & & \multicolumn{3}{c}{HGS-SSP}\\
\cline{5-7}
\cline{9-11}
\cline{13-15}
$n$ & $m$ & $C$ & $i$ & Best & Avg & T & & Best & Avg & T & & Best & Avg & T \\
\hline
10 & 10 & 4 & 10 & \bftab{9.50} & \bftab{9.50} & 1.70 & & \bftab{9.50} & \bftab{9.50} & 0.08 & & \bftab{9.50} & \bftab{9.50} & 0.07 \\
10 & 10 & 5 & 10 & \bftab{6.20} & 6.21 & 2.15 & & \bftab{6.20} & \bftab{6.20} & 0.08 & & \bftab{6.20} & \bftab{6.20} & 0.06 \\
10 & 10 & 6 & 10 & \bftab{4.30} & \bftab{4.30} & 3.16 & & \bftab{4.30} & \bftab{4.30} & 0.05 & & \bftab{4.30} & \bftab{4.30} & 0.06 \\
10 & 10 & 7 & 10 & \bftab{3.00} & \bftab{3.00} & 3.79 & & \bftab{3.00} & \bftab{3.00} & 0.03 & & \bftab{3.00} & \bftab{3.00} & 0.06 \\
15 & 20 & 6 & 10 & \bftab{21.40} & 21.71 & 7.79 & & \bftab{21.40} & \bftab{21.40} & 0.75 & & \bftab{21.40} & \bftab{21.40} & 0.92 \\
15 & 20 & 8 & 10 & \bftab{14.20} & 14.33 & 8.43 & & \bftab{14.20} & \bftab{14.20} & 0.87 & & \bftab{14.20} & 14.21 & 0.84 \\
15 & 20 & 10 & 10 & \bftab{10.30} & 10.34 & 13.96 & & \bftab{10.30} & \bftab{10.30} & 0.67 & & \bftab{10.30} & \bftab{10.30} & 0.72 \\
15 & 20 & 12 & 10 & \bftab{8.20} & \bftab{8.20} & 16.44 & & \bftab{8.20} & \bftab{8.20} & 0.60 & & \bftab{8.20} & \bftab{8.20} & 0.69 \\
 \hline
\end{tabular}
\label{tab:groupE}
\end{table}

On the first five groups of instances, we observe that HGS-SSP finds, in general, solutions of similar quality as ILS. Group $A$ contains very small instances, and thus all methods find the BKSs. For group $B$, our algorithm obtained the BKS for all instances, with a CPU time slightly smaller than that of ILS, and much faster than that of CS+BRKGA. For groups $C$, $D$ and $E$, HGS-SSP found in most cases the same solution as ILS with similar CPU times. Since these instances are relatively small, however, very little differences between methods can be generally observed, and we should turn towards larger problem instances with better discriminating power.

Tables \ref{tab:groupCrama} and \ref{tab:groupdat} now compare the performance of existing methods on the larger instances of \citet{Crama1994} and \citet{Catanzaro2015}. On these instances, very significant differences of performance can be noticed.

\begin{table}[htbp!]
\centering
\renewcommand{\arraystretch}{1.2}
\setlength{\tabcolsep}{3.5pt}
\caption{Groups $C_1$, $C_2$, $C_3$, and $C_4$}
\footnotesize{}
\begin{tabular}{ l l l l l l l l l l l l l l l }
\hline
& & & & \multicolumn{3}{c}{ILS} & & \multicolumn{3}{c}{DQGA} & & \multicolumn{3}{c}{HGS-SSP}\\
\cline{5-7}
\cline{9-11}
\cline{13-15}
$n$ & $m$ & $C$ & $i$ & Best & Avg & T & & Best & Avg & T & & Best & Avg & T \\
\hline
10 & 10 & 4 & 10 & \bftab{9.10} & \bftab{9.10} & 0.08 & & \bftab{9.10} & \bftab{9.10} & 0.23 & & \bftab{9.10} & \bftab{9.10} & 0.08 \\
10 & 10 & 5 & 10 & \bftab{6.20} & \bftab{6.20} & 0.08 & & \bftab{6.20} & \bftab{6.20} & 0.26 & & \bftab{6.20} & \bftab{6.20} & 0.07 \\
10 & 10 & 6 & 10 & \bftab{4.30} & \bftab{4.30} & 0.06 & & \bftab{4.30} & \bftab{4.30} & 0.29 & & \bftab{4.30} & \bftab{4.30} & 0.06 \\
10 & 10 & 7 & 10 & \bftab{3.10} & \bftab{3.10} & 0.04 & & \bftab{3.10} & \bftab{3.10} & 0.25 & & \bftab{3.10} & \bftab{3.10} & 0.06 \\
15 & 20 & 6 & 10 & \bftab{20.60} & 20.63 & 0.82 & & \bftab{20.60} & 20.70 & 2.27 & & \bftab{20.60} & \bftab{20.60} & 0.93 \\
15 & 20 & 8 & 10 & \bftab{13.70} & \bftab{13.70} & 1.06 & & \bftab{13.70} & \bftab{13.70} & 1.55 & & \bftab{13.70} & \bftab{13.70} & 0.82 \\
15 & 20 & 10 & 10 & \bftab{10.10} & \bftab{10.10} & 0.84 & & \bftab{10.10} & \bftab{10.10} & 1.07 & & \bftab{10.10} & \bftab{10.10} & 0.73 \\
15 & 20 & 12 & 10 & \bftab{7.60} & \bftab{7.60} & 0.58 & & \bftab{7.60} & \bftab{7.60} & 0.64 & & \bftab{7.60} & \bftab{7.60} & 0.64 \\
30 & 40 & 15 & 10 & 91.40 & 91.70 & 84.21 & & 91.30 & 91.88 & 184.54 & & \bftab{91.10} & \bftab{91.10} & 75.85 \\
30 & 40 & 17 & 10 & 71.30 & 71.59 & 128.05 & & \bftab{71.20} & 72.08 & 254.18 & & \bftab{71.20} & \bftab{71.20} & 66.94 \\
30 & 40 & 20 & 10 & 50.40 & 50.71 & 168.27 & & 50.40 & 51.36 & 224.38 & & \bftab{50.20} & 50.37 & 64.40 \\
30 & 40 & 25 & 10 & 27.90 & 28.14 & 142.41 & & \bftab{27.70} & 28.64 & 189.77 & & 27.80 & 28.02 & 62.61 \\
40 & 60 & 20 & 10 & 178.40 & 179.00 & 428.41 & & 178.40 & 180.92 & 599.65 & & \bftab{177.20} & 177.41 & 512.09 \\
40 & 60 & 22 & 10 & 151.50 & 152.04 & 645.19 & & 151.30 & 154.00 & 792.20 & & \bftab{150.50} & 150.67 & 490.88 \\
40 & 60 & 25 & 10 & 121.00 & 121.52 & 985.88 & & 120.90 & 123.90 & 979.43 & & \bftab{120.20} & 120.44 & 478.33 \\
40 & 60 & 30 & 10 & 83.90 & 84.18 & 1440.42 & & 83.90 & 86.30 & 1388.76 & & \bftab{83.20} & 83.44 & 408.90 \\
 \hline
\end{tabular}
\label{tab:groupCrama}
\end{table}

\begin{table}[htbp!]
\centering
\renewcommand{\arraystretch}{1.2}
\setlength{\tabcolsep}{3.5pt}
\caption{Groups $datA$, $datB$, $datC$ and $datD$}
\footnotesize{}
\begin{tabular}{ l l l l l l l l l l l l l l l }
\hline
& & & & \multicolumn{3}{c}{ILS} & & \multicolumn{3}{c}{DQGA} & & \multicolumn{3}{c}{HGS-SSP}\\
\cline{5-7}
\cline{9-11}
\cline{13-15}
$n$ & $m$ & $C$ & $i$ & Best & Avg & T & & Best & Avg & T & & Best & Avg & T \\
\hline
10 & 10 & 4 & 10 & \bftab{8.50} & \bftab{8.50} & 0.09 & & \bftab{8.50} & \bftab{8.50} & 0.19 & & \bftab{8.50} & \bftab{8.50} & 0.07 \\
10 & 10 & 5 & 10 & \bftab{5.80} & \bftab{5.80} & 0.09 & & \bftab{5.80} & \bftab{5.80} & 0.19 & & \bftab{5.80} & \bftab{5.80} & 0.06 \\
10 & 10 & 6 & 10 & \bftab{4.10} & \bftab{4.10} & 0.05 & & \bftab{4.10} & \bftab{4.10} & 0.22 & & \bftab{4.10} & \bftab{4.10} & 0.06 \\
10 & 10 & 7 & 10 & \bftab{3.00} & \bftab{3.00} & 0.04 & & \bftab{3.00} & \bftab{3.00} & 0.25 & & \bftab{3.00} & \bftab{3.00} & 0.05 \\
15 & 20 & 6 & 10 & \bftab{20.50} & \bftab{20.50} & 1.06 & & \bftab{20.50} & 20.72 & 1.85 & & \bftab{20.50} & \bftab{20.50} & 0.88 \\
15 & 20 & 8 & 10 & \bftab{13.70} & \bftab{13.70} & 1.34 & & \bftab{13.70} & 13.74 & 2.42 & & \bftab{13.70} & \bftab{13.70} & 0.79 \\
15 & 20 & 10 & 10 & \bftab{9.70} & \bftab{9.70} & 1.09 & & \bftab{9.70} & 9.76 & 1.58 & & \bftab{9.70} & \bftab{9.70} & 0.69 \\
15 & 20 & 12 & 10 & \bftab{7.20} & \bftab{7.20} & 0.69 & & \bftab{7.20} & \bftab{7.20} & 0.90 & & \bftab{7.20} & \bftab{7.20} & 0.60 \\
30 & 40 & 15 & 10 & 83.90 & 84.09 & 96.49 & & 83.80 & 84.82 & 165.49 & & \bftab{83.50} & \bftab{83.50} & 74.71 \\
30 & 40 & 17 & 10 & 65.50 & 65.82 & 132.97 & & 65.50 & 66.74 & 223.35 & & \bftab{65.40} & 65.43 & 71.77 \\
30 & 40 & 20 & 10 & \bftab{46.60} & 46.78 & 167.41 & & \bftab{46.60} & 47.82 & 250.95 & & \bftab{46.60} & 46.66 & 60.58 \\
30 & 40 & 25 & 10 & 26.30 & 26.47 & 133.27 & & \bftab{26.20} & 27.10 & 165.03 & & \bftab{26.20} & 26.32 & 48.01 \\
40 & 60 & 20 & 10 & 178.00 & 178.36 & 474.00 & & 178.00 & 180.16 & 243.10 & & \bftab{176.50} & 176.71 & 501.21 \\
40 & 60 & 22 & 10 & 151.60 & 152.05 & 684.95 & & 151.50 & 153.68 & 552.95 & & \bftab{150.30} & 150.45 & 437.85 \\
40 & 60 & 25 & 10 & 121.20 & 121.68 & 1037.20 & & 121.20 & 123.68 & 853.43 & & \bftab{120.30} & 120.61 & 466.12 \\
40 & 60 & 30 & 10 & 85.20 & 85.42 & 1473.24 & & 85.20 & 87.22 & 1296.50 & & \bftab{84.30} & 84.66 & 484.14 \\
 \hline
\end{tabular}
\label{tab:groupdat}
\end{table}

For the first two groups of each table ($C_1$ and $C_2$ from Table \ref{tab:groupCrama} and $datA$ and $datB$ from Table \ref{tab:groupdat}), detailed in the first eight lines, the average solutions of HGS-SSP match the BKS on all instances. HGS-SSP consumes a CPU time similar to that of ILS, and much shorter than that of DQGA. For the last two groups of each table ($C_3$ and $C_4$ from Table \ref{tab:groupCrama} and $datC$ and $datD$ from Table \ref{tab:groupdat}), HGS-SSP finds the same or new BKSs, with only one exception, and produces significantly better average results than all other methods for a CPU which is two to three times smaller. These significant improvements, on larger instances, show that the proposed method performs a more sustained and diversified search in the solution space.\\

\noindent
\textbf{Comparison with known optimal solutions.}
Finally, some optimal solutions are known for those instances from previous studies. In particular, \citet{Yanasse2009} solved the small instances of groups $A$, $B$ and $E$ to optimality, as well as some instances of groups $C$ and $D$. Moreover, \citet{Catanzaro2015} solved to optimality some instances with $10$ jobs and $10$ tools from group $datA$. For all these instances, we verified that the proposed HGS-SSP retrieved the optimal solutions on every run.

\subsection{Experiments on larger instances}

The benchmark instances of \citet{Crama1994}, \citet{Yanasse2009}, and \citet{Catanzaro2015} have been widely used in the literature. However, these instances remain limited in terms of number of jobs and tools, and start to lose their discriminative power: state-of-the-art heuristics now find solutions which are close to each other. Moreover, as discussed in \citet{Shirazi2001}, companies are regularly confronted with problems that contain over sixty jobs. To stimulate future research on the SSP and allow comparisons on more challenging instances, we generated a set of additional instances of a size comparable to that reported in \citet{Shirazi2001}, and provide in Table \ref{tab:grouplarge} (in Appendix) the results of our method for future reference. These datasets are also available at \url{https://w1.cirrelt.ca/~vidalt/en/research-data.html}.

The new instances can be divided into three groups ($F_1, F_2$ and $F_3$), each with the same number of jobs and tools, but different capacities. For each capacity, we generated five instances (each line from \autoref{tab:grouplarge}). The groups have a number of jobs ranging from 50 to 70 and a number of tools ranging from 75 to 105. Despite the higher CPU time required to solve these larger instances, the objective values obtained by HGS-SSP over multiple runs remained very stable. This shows that the method is robust in the sense that it consistently achieves a similar solution quality. Future research on these instances will allow more extensive algorithm comparisons.

\section{Conclusions}
\label{sec:conclusion}

In this article, we proposed a simple and efficient metaheuristic for the well-known job sequencing and tool switching problem (SSP). The proposed approach contrasts with previous algorithms, which had a tendency to be over-engineered and generally complex. Our HGS-SSP finds a good balance between aggressive \emph{intensification}, achieved by an efficient local search in the space of permutations, and \emph{diversification} \citep{Blum2003,Vidal2012a}, obtained via a simple crossover, population diversity management strategy, and tie-breaking objective to guide the search towards potential tool switches reductions. Through extensive experiments on several sets of benchmark instances, we observed that our algorithm significantly outperforms existing methods in terms of solution quality and CPU time. To guide future research, we also evaluated the impact of each neighborhood and methodological choice in the method. We observed that adopting a defined order of neighborhood exploration in the local search is beneficial for this problem since \textsc{2opt} tends to operate larger structural changes than \textsc{Swap} or \textsc{Relocate}. The tie-breaking objective was also essential for a good performance, along with our population diversity management techniques.

Many promising research perspectives are open. The proposed HGS-SSP could be extended and tested on other SSP variants, e.g., with multiple machines \citep{Beezao2017} or with different objective functions \citep{Camels2019}.
As the proposed method exploits an indirect solution representation as a job permutation with a solution decoder (the KTNS algorithm), it conducts the search on a much smaller space at the price of more computationally expensive solution evaluations. Such a disciplined analysis of metaheuristics and structural problem decompositions should be pursued and extended to other problems, seeking to achieve a search space reduction which is as large as possible for a decoder which is a fast as possible, opening the way to a variety of complexity analyses and algorithmic contributions. Finally, the HGS framework could be extended to efficiently solve other difficult permutation-based optimization problems.

\section*{Acknowledgments}
This research was partially supported by Conselho Nacional de Desenvolvimento Cient{\'i}fico e Tecnol{\'o}gico (CNPq), grants \mbox{428549/2016-0}, \mbox{307843/2018-1} and \mbox{308528/2018-2}, by FAPERJ (grant \mbox{E-26/203.310/2016}), and by Comiss\~{a}o de Aperfei\c{c}oamento de Pessoal de N{\'i}vel Superior (CAPES) -- Finance Code 001.


\appendixtitleon
\appendixtitletocon
\begin{appendices}
\section{Results on a the new set of larger instances}

\begin{table}[htbp!]
\centering
\renewcommand{\arraystretch}{1.2}
\caption{Groups $F_1$, $F_2$ and $F_3$}
\footnotesize{}
\begin{tabular}{ l l l l l l l }
\hline
& & & & \multicolumn{3}{c}{HGS-SSP}\\
\cline{5-7}
$n$ & $m$ & $C$ & $i$ & Best & Avg & T \\
\hline
50 & 75 & 25 & 5 & 268.40 & 268.80 & 2349.09  \\
50 & 75 & 30 & 5 & 196.60 & 197.22 & 2064.81 \\
50 & 75 & 35 & 5 & 147.00 & 148.14 & 1848.54 \\
50 & 75 & 40 & 5 & 109.80 & 110.62 & 1607.07 \\
60 & 90 & 35 & 5 & 414.80 & 415.54 & 7607.10 \\
60 & 90 & 40 & 5 & 320.20 & 321.08 & 7108.39 \\
60 & 90 & 45 & 5 & 247.00 & 248.34 & 5253.31 \\
60 & 90 & 50 & 5 & 191.20 & 192.86 & 5595.93 \\
70 & 105 & 40 & 5 & 576.80 & 577.82 & 16353.88 \\
70 & 105 & 45 & 5 & 459.00 & 460.18 & 14264.06 \\
70 & 105 & 50 & 5 & 369.20 & 371.00 & 12360.99 \\
70 & 105 & 55 & 5 & 298.20 & 299.80 & 11756.43 \\
 \hline
\end{tabular}
\label{tab:grouplarge}
\end{table}

\end{appendices}

\end{document}